\documentclass[letterpaper, 10 pt, conference]{ieeeconf}
\IEEEoverridecommandlockouts
\usepackage[utf8]{inputenc}
\usepackage{amsfonts}
\usepackage{float}
\usepackage[T1]{fontenc}
\usepackage[noadjust]{cite}

\usepackage[english]{babel}
\usepackage [autostyle, english = american]{csquotes}
\MakeOuterQuote{"}
\usepackage{textcomp}
\usepackage{blindtext}
\usepackage{graphicx}
\usepackage{lettrine}
\usepackage{graphicx}
\usepackage{epstopdf}
\usepackage{lipsum}
\usepackage{multicol}
\usepackage{multirow}
\usepackage{textcomp}
\let\labelindent\relax
\usepackage{enumitem}
\usepackage[font=small,skip=0pt]{caption}
\usepackage{amsmath}
\usepackage{color}
\usepackage{mathtools}
\usepackage{booktabs}

\usepackage{array}
\newcolumntype{L}[1]{>{\raggedright\let\newline\\\arraybackslash\hspace{0pt}}m{#1}}
\newcolumntype{C}[1]{>{\centering\let\newline\\\arraybackslash\hspace{0pt}}m{#1}}
\newcolumntype{R}[1]{>{\raggedleft\let\newline\\\arraybackslash\hspace{0pt}}m{#1}}
\usepackage[linesnumbered]{algorithm2e}
\usepackage{bbm}
\usepackage{duckuments}
\usepackage{hyperref}
\hypersetup{colorlinks,linkcolor={black},citecolor={blue},urlcolor={blue}}
\usepackage[noadjust]{cite}

\DeclareMathOperator{\sech}{sech}

\newcolumntype{P}[1]{>{\centering\arraybackslash}p{#1}}

\title{Quaternion-Based Sliding Mode Control for Six Degrees of Freedom Flight Control of Quadrotors}

\author{Amin Yazdanshenas, and Reza Faieghi
\thanks{The authors are with the Autonomous Vehicles Laboratory, Department of Aerospace Engineering, Toronto Metropolitan University, Toronto, Canada{\tt\footnotesize \{amin.yazdanshenas,reza.faieghi\}@torontomu.ca}. This work was partially supported by the Natural Sciences and Engineering Research Council (NSERC) of Canada.}
}
\begin{document}
\setlength{\intextsep}{1pt plus 1pt minus 1pt} 
\setlength{\textfloatsep}{1pt plus 1pt minus 1pt} 

\maketitle
\thispagestyle{empty}
\pagestyle{empty}
\bstctlcite{IEEEexample:BSTcontrol}
\markboth{IEEE Robotics and Automation Letters}
{Yazdanshenas, and Faieghi: Sliding Mode Control for Robust Trajectory Tracking of Quadrotors}

\begin{abstract}
Despite extensive research on sliding mode control (SMC) design for quadrotors, the existing approaches have certain limitations.
Euler angle-based SMC formulations suffer from poor performance in high-pitch or -roll maneuvers.
Quaternion-based SMC approaches have unwinding issues and complex architecture. 
Coordinate-free methods are slow and only almost globally stable.
This paper presents a new six degrees of freedom SMC flight controller to address the above limitations.
We use a cascaded architecture with a position controller in the outer loop and a quaternion-based attitude controller in the inner loop.
The position controller generates the desired trajectory for the attitude controller using a coordinate-free approach.
The quaternion-based attitude controller uses the natural characteristics of the quaternion hypersphere, featuring a simple structure while providing global stability and avoiding unwinding issues.
We compare our controller with three other common control methods conducting challenging maneuvers like flip-over and high-speed trajectory tracking in the presence of model uncertainties and disturbances.
Our controller consistently outperforms the benchmark approaches with less control effort and actuator saturation, offering highly effective and efficient flight control.
\end{abstract}



\textit{Notations:} Scalars are in italics, vectors are in lowercase bold, and matrices are in uppercase bold.

\section{Introduction}\label{sec:Intro}
\lettrine{Q}{uadrotors} require a robust controller to maintain stability and precise and agile maneuverability in the face of model uncertainties and external disturbances.
One common control methodology that comes to mind when discussing robust control is sliding mode control (SMC), known for its simplicity and robustness \cite{slotine1991applied}.
SMC is proven effective in many application areas \cite{DANGELO2024104554,nemati2019fast,10156565, faieghi2012novel}; however, its current implementations for quadrotors have certain shortcomings.

Previous work on sliding mode controller design for quadrotors can be categorized based on their formulation in modeling the vehicle rotational dynamics.
An ample body of literature relies on the Euler angle representation of the vehicle attitude \cite{xu2006sliding, zheng2014second, xiong2017global, izadi2024high, thanh2018quadcopter}.
While robust 6-DOF flight control can be achieved with such methods, they are only suitable for small roll- or pitch-angle flight.

To elaborate, let us first establish the attitude dynamics of a quadrotor using Euler angles. 
Let ${\boldsymbol{\eta}}=\left[\phi, \theta,\psi\right]^{\top}$, where $ - \pi  < \phi  \le \pi $, $ - \frac{\pi }{2} \le \theta  \le \frac{\pi }{2}$, and $ - \pi  < \psi  \le \pi $ be the Euler angles representing pitch, roll, and yaw in the yaw-pitch-roll sequence,  ${\boldsymbol{\omega}}$ the angular velocity vector, ${\rm{\mathbf{J}}}=\operatorname{diag}\left(\left[J_x, J_y, J_z\right] \right)$ the inertia matrix, and $\boldsymbol{\tau}$ the moments around the principal axes.
The attitude dynamics becomes
\begin{equation}\label{eq:EulerKinematicalEquation}
    \dot {\boldsymbol{\eta}} = {\bf{H}}\left({\boldsymbol{\eta}}\right) {\boldsymbol{\omega}},
\end{equation}
where  
\begin{equation}\label{eq:H}
{\bf{H}}\left( {\boldsymbol{\eta }} \right) = \left[ {\begin{array}{*{20}{c}}
1&{\sin \phi \tan \theta }&{\cos \phi \tan \theta }\\
0&{\cos \phi }&{ - \sin \phi }\\
0&{{{\sin \phi } \mathord{\left/
 {\vphantom {{\sin \phi } {\cos \theta }}} \right.
 \kern-\nulldelimiterspace} {\cos \theta }}}&{{{\cos \phi } \mathord{\left/
 {\vphantom {{\cos \phi } {\cos \theta }}} \right.
 \kern-\nulldelimiterspace} {\cos \theta }}}
\end{array}} \right],
\end{equation}
and
\begin{equation}\label{eq:rotDyn}
    \dot{\boldsymbol{\omega}} = {\bf{J}}^{ - 1}\left( -{\boldsymbol{\omega}} \times {\bf{J}}{\boldsymbol{\omega}}+ \boldsymbol{\tau} \right),
\end{equation}
where $\times$ indicates the cross product.
Now, if $\phi$ and $\theta$ are small, $\bf{H}$ can be greatly simplified such that $\dot{\boldsymbol{\eta}} \approx {\boldsymbol{\omega}}$. Subsequently, \eqref{eq:EulerKinematicalEquation}-\eqref{eq:rotDyn} can be simplified to a set of second-order differential equations
\begin{equation}
\left\{ {\begin{array}{*{20}{l}}
{\ddot \phi  = \dot \theta \dot \psi (\frac{{{J_y} - {J_z}}}{{{J_x}}})   + \frac{1}{{{J_x}}}{\tau_1}},\\
{\ddot \theta  = \dot \varphi \dot \psi (\frac{{{J_z} - {J_x}}}{{{J_y}}})   + \frac{1}{{{J_y}}}{\tau_2}},\\
{\ddot \psi  = \dot \varphi \dot \theta (\frac{{{J_x} - {J_y}}}{{{J_z}}}) + \frac{1}{{{J_z}}}{\tau_3}}.\\
\end{array}} \right.
\end{equation}
The above equations have been the basis of numerous SMC designs for quadrotors \cite{xu2006sliding, zheng2014second, xiong2017global, izadi2024high, thanh2018quadcopter}. 
While their particular second-order structure lends well to the SMC formulations, they are only valid for small $\phi$ and $\theta$ values. 
Thus, as we will show in our results, the controllers using such models struggle with maneuvers that entail large pitch and roll angles.

Another category of related papers relies on a quaternion-based representation of attitude.
Several papers in this category lack 6-DOF SMC designs, applying SMC for only position control or attitude control \cite{esmail2022attitude,abaunza2019quadrotor}.
Several other papers have explored 6-DOF SMC designs \cite{serrano2023terminal, arellano2015quaternion,sanwale2020quaternion}; however, they suffer from one or two of the following limitations:
(i) the control architecture is complex, and (ii) there is the quaternion unwinding problem that can lead to longer-than-necessary attitude maneuvers.

The third category of papers uses the geometric control approaches, designing SMC in $\mathrm{SE}(3)$ \cite{garcia2020robust,gong2020adaptive,ren2023adaptive}. 
In this approach, the attitude-tracking error is defined as ${\mathbf{R}}_e = {\mathbf{R}}^{\top}_d \mathbf{R}$, where $\mathbf{R}$ is the rotation matrix, and the indices $e$ and $d$ represent the error and desired value \cite{lee2010geometric}.
This method is prone to slow convergence \cite{teng2022lie, lopez2020sliding} and leads to almost global stability \cite{bhat2000topological}.

In this paper, we address the above challenges by developing a new quaternion-based 6-DOF sliding mode controller, avoiding the aforementioned limitations of Euler-based and $\mathrm{SE}(3)$  approaches. We adopt the method presented in \cite{lopez2020sliding} to eliminate the unwinding problem of quaternions. We apply a cascaded control design, where the position controller output in the outer loop is used to generate the desired trajectory for the attitude controller. We use a coordinate-free desired attitude generation method \cite{lee2010geometric} with no simplification of rotational dynamics, enabling effective flight control during aggressive maneuvers in the presence of model uncertainties and external disturbances. Our controller features global stability and uses the inherent characteristics of quaternion dynamics in $\mathbb{S}^3$ to achieve such remarkable performance.
We compare our results with common controllers \cite{lee2010geometric, fresk2013full, bhat2000topological}, and show consistent improvements in flight performance while minimizing actuator effort and saturation.

\section{Quadrotor dynamics}\label{sec:Dynamics}
This section presents the equations of flight for quadrotors to establish the notation for our control developments.

\begin{figure}
    \centering
    \includegraphics[trim=6cm 0cm 7cm 1.4cm,clip,width=0.8\linewidth]{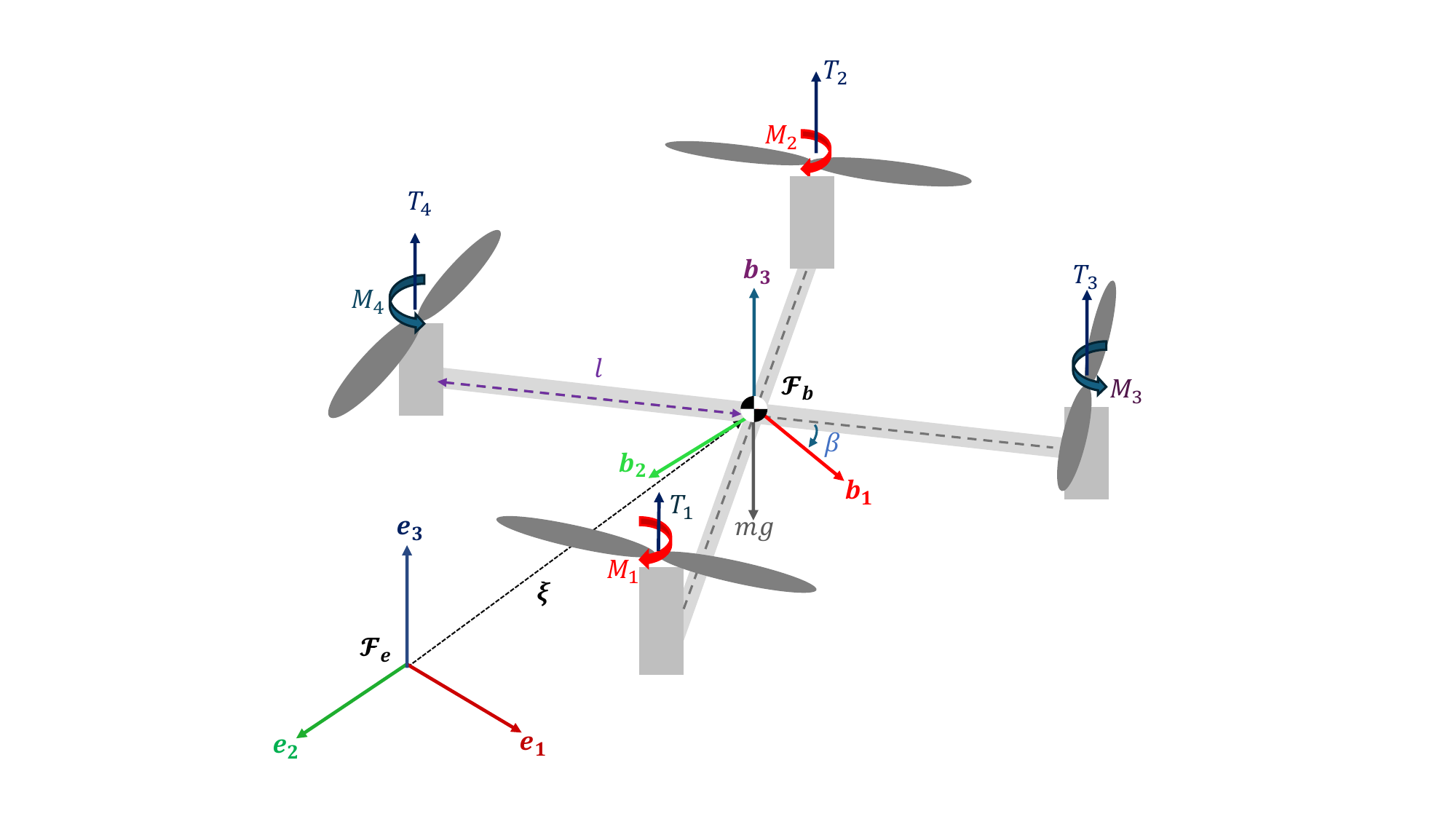}
    \caption{Quadrotor model and coordinate frames.}
    \label{fig:quadrotor_mdoel}
\end{figure}

Consider the inertial and body-fixed frames, $\mathcal{F}_e = \left\{\mathbf{e}_1,\mathbf{e}_2,\mathbf{e}_3\right\}$ and  $\mathcal{F}_b=\left\{\mathbf{b}_1,\mathbf{b}_2,\mathbf{b}_3\right\}$, as illustrated in Fig. \ref{fig:quadrotor_mdoel}. 
Define $\boldsymbol{\xi} = \left[x, y, z\right]^{\top}$ to be the position of the vehicle, $\mathbf{v} \in \mathbb{R}^3$ the velocity, $\mathbf{q} = \left[q_{w}, \Vec{\mathbf{q}}^{\top} \right]^{\top} \in \mathbb{S}^3$ the unit quaternion representing the attitude, and $\boldsymbol{\omega} \in \mathbb{R}^3$ the angular velocity. 
Then, the quadrotor flight dynamics is 
\begin{equation}\label{eq:dynamic}
    \dot{\mathbf{x}} = 
    \begin{bmatrix}
        \dot {\boldsymbol{\xi}} \\ \dot{\mathbf{v}} \\ \dot{\mathbf{q}} \\ \dot{\boldsymbol{\omega}}
    \end{bmatrix} =
    \begin{bmatrix}
        \mathbf{v} \\  -g\mathbf{e}_{3} + \frac{1}{m} \mathbf{q} \otimes
            f\mathbf{e}_{3}
 \otimes \mathbf{q}^* + \mathbf{d}_a \\
        \frac{1}{2}\mathbf{q} \otimes
        \begin{bmatrix}
            0 \\ \boldsymbol{\omega}
        \end{bmatrix}
    \\ \mathbf{J}^{-1} \left(-\boldsymbol{\omega} \times \mathbf{J}\boldsymbol{\omega}  + \boldsymbol{\tau} \right) + \mathbf{d}_{\alpha}
    \end{bmatrix},
\end{equation}
where $\otimes$ is the quaternion product, $\mathbf{q}^*$ is the conjugate of $\mathbf{q}$, $g$ is the gravity, $m$ is the mass, $\mathbf{J}$ is the inertia matrix, $f$ is the thrust, $\boldsymbol{\tau}$ is the moments, and $\mathbf{d}_a$ and $\mathbf{d}_\alpha$ are bounded external disturbances.

We define the control input $\mathbf{u} \in \mathbb{R}^4$ to be the vector of thrusts generated by the rotors, given by $\mathbf{u} = c_t\boldsymbol{\Omega}^{\circ2}$,
where $\mathbf{\Omega}$ is the vector encompassing the angular rate of rotors, $c_t$ is the rotors' thrust coefficient, and $^{\circ}$ is the Hadamard power.
The control input maps to the force and moments applied on the quadrotor through
\begin{equation}
    \begin{bmatrix}
        f\\ \boldsymbol{\tau}
    \end{bmatrix}
    = \mathbf{G}\mathbf{u},
\end{equation}
where
\begin{equation}
    \mathbf{G} = \begin{bmatrix}
        1&1&1&1\\
        -l\;\sin(\beta) & l\;\sin(\beta)&l\;\sin(\beta) & -l\;\sin(\beta)\\
        l\;\cos(\beta) & -l\;\cos(\beta)&l\;\cos(\beta) & -l\;\cos(\beta)\\
        -c_q/c_t &-c_q/c_t &c_q/c_t &c_q/c_t 
    \end{bmatrix}.
\end{equation}
with $c_q$ being the rotors' torque coefficient and $\beta$ and $l$ being geometric parameters defined in Fig. \ref{fig:quadrotor_mdoel}.

\section{Controller Design}\label{sec:Control}
This section presents the proposed 6-DOF quaternion-based sliding mode controller for quadrotors.
\begin{figure*}[t]
    \centering
    \includegraphics[trim=0cm 2cm 0cm 0.3cm,clip,width=0.75\linewidth]{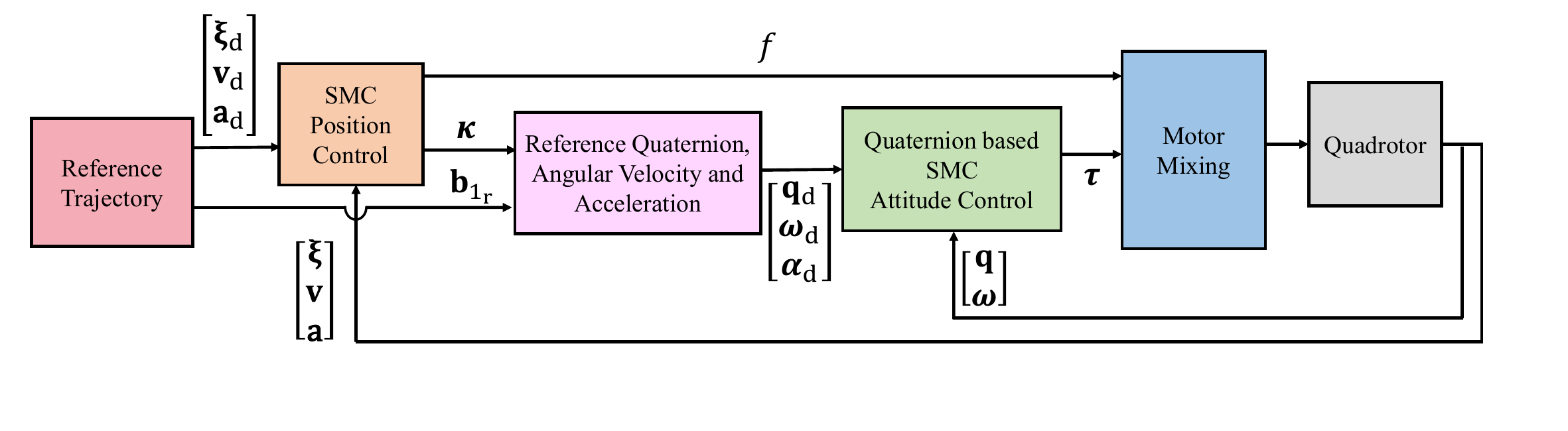}
    \caption{Controller architecture.}
    \label{fig:cascaded}
\end{figure*}
Figure \ref{fig:cascaded} illustrates the controller architecture.
The reference trajectory consist of desired position $\boldsymbol{\xi}_d$ and heading ${\mathbf{b}}_{1_r}= \left[\cos{\psi_d}, \sin{\psi_d}, 0\right]^{\top}$.
To deal with the vehicle under-actuation, we employ a cascaded control structure including position and attitude control loops.
We use SMC for both position and attitude control, enabling the robustness benefits of SMC for both control loops, as opposed to many position-only or attitude-only SMC designs.
The position controller determines the desired thrust and the desired attitude.
However, contrary to existing 6-DOF SMC designs that rely on Euler angles, we employ a coordinate-free approach to generate the desired attitude.
For the attitude controller itself, we use a quaternion-based SMC.
This approach leads to a new 6-DOF SMC with high effectiveness and efficiency.

\subsection{Position controller}\label{subsec:posControl}
Let $\boldsymbol{{\xi}}_{d}$, $\mathbf{v}_d$, and $\mathbf{a}_d$ be the desired position, velocity, and acceleration. Then, the error terms take the following forms
\begin{equation}\label{eq:postionErrorDyn}
\begin{gathered}
    {\boldsymbol{\xi}_e  =   \boldsymbol{\xi} -\boldsymbol{\xi}_d}\;,
    {\mathbf{v}_e  =   \mathbf{{v}} -\mathbf{{v}}_d}, \;\text{and}\;\\
    {\mathbf{a}_e  =   \dot{\mathbf{{v}}} -\dot{\mathbf{{v}}}_d} = -{g\mathbf{e}_{3}} + \frac{1}{m} \mathbf{q} \otimes
            f\mathbf{e}_{3}
         \otimes \mathbf{q}^*  - \ddot{\boldsymbol{\xi}}_d.
\end{gathered}  
\end{equation}
Using the standard SMC formulation, we define the position controller sliding surface as 
\begin{equation}
    \mathbf{s_{\xi}}  = \mathbf{v}_{e} + \operatorname{diag}(\boldsymbol{\lambda}_{\xi}^{\top})\boldsymbol{\xi}_{e},
\end{equation}
where $\boldsymbol{\lambda}_{\xi}$ is a design parameter with positive elements.
To ensure chattering-free control, instead of using the well-known sliding condition $\dot{\mathbf{s}}_{\xi} = -\mathbf{K}_{\xi}{\rm{sgn}}\left(\mathbf{s}_{\xi}\right)$, we use
\begin{equation}\label{eq:slidingCondition}
    \dot{\mathbf{s}}_{\xi}  =  -{\mathbf{K}}_{\xi}\tanh(\mathbf{s}_{\xi}),
\end{equation}
where ${\mathbf{K}}_{\xi}$ is a design parameter.
Substituting $\dot{\mathbf{s}}_{\xi}  =   \mathbf{a}_e + \operatorname{diag}(\boldsymbol{\lambda}_{\xi}^{\top}) \mathbf{v}_{e}$ in \eqref{eq:slidingCondition} and \eqref{eq:postionErrorDyn} leads to
\begin{equation}\label{eq:posControlDerivations}
\begin{split}
\operatorname{diag}(\boldsymbol{\lambda}_{\xi}^{\top}) \mathbf{v}_{e} -{g\mathbf{e}_{3}} + \frac{1}{m} \mathbf{q} \otimes f\mathbf{e}_{3}
         \otimes \mathbf{q}^*  - \ddot{\boldsymbol{\xi}}_d \\=  -\mathbf{K}_{\xi}\tanh(\mathbf{s_{\xi}}).
        \end{split}
\end{equation}
Note that using quaternion multiplication, we have
\begin{equation}\label{eq:posControlDerivations2}
       \mathbf{q} \otimes 
            f\mathbf{e}_{3}
        \otimes \mathbf{q}^*=f\begin{bmatrix}
            2(q_x q_z + q_w q_y)\\
            2(q_y q_z - q_w q_x)\\
q_w^2 - q_x^2 - q_y^2 + q_z^2\\
        \end{bmatrix}=\boldsymbol{\kappa},
\end{equation}

where
\begin{equation}\label{eq:A}
    {\boldsymbol{\kappa}} =  {m \left( \ddot{\boldsymbol{\xi}}_d -\operatorname{diag}(\boldsymbol{\lambda}_{\xi}^{\top}) \mathbf{v}_{e} + {g\mathbf{e}_{3}} -\mathbf{K}_{\xi}\tanh(\mathbf{s}_{\xi})\right)}.
\end{equation}
It follows that 
\begin{equation}
        f= \boldsymbol{\kappa} \cdot \begin{bmatrix}
            2(q_x q_z + q_w q_y)\\
            2(q_y q_z - q_w q_x)\\
q_w^2 - q_x^2 - q_y^2 + q_z^2\\
        \end{bmatrix},
\end{equation}
which constitutes our control law for thrust.
\subsection{Trajectory generation for the attitude controller}
The auxiliary variable $\boldsymbol{\kappa}$ can be used to generate the desired quaternion ${\mathbf{q}}_d$, desired angular velocity $\boldsymbol{\omega}_d$, and desired angular acceleration $\boldsymbol{\alpha}_d$ for the attitude controller. 
Inspired by the method in \cite{lee2010geometric}, we consider the desired rotation matrix $
    \mathbf{R}_{d} = \left[\mathbf{b}_{1_d}, \mathbf{b}_{2_d}, \mathbf{b}_{3_d}\right]$,
where $\mathbf{b}_{3_d} = {\boldsymbol{\kappa}}/{\lVert \boldsymbol{\kappa} \rVert}$.
The goal is to find $\mathbf{b}_{1_d}$ and $\mathbf{b}_{2_d}$ to construct ${\mathbf{R}}_{d}$, which will lead to finding ${\mathbf{q}}_d$, $\boldsymbol{\omega}_d$, and $\boldsymbol{\alpha}_d$.

Starting with the reference heading $\mathbf{b}_{1_r}$ and $\mathbf{b}_{3_d}$ defined above, we define $\mathbf{b}_{2_d}$ as $\mathbf{b}_{2_d} = {\boldsymbol{\nu}}/{\lVert \boldsymbol{\nu} \rVert}$ where $\boldsymbol{\nu} = \mathbf{b}_{3_d} \times \mathbf{b}_{1_r}$.
Next, we define $\mathbf{b}_{1_d} = \mathbf{b}_{2_d} \times \mathbf{b}_{3_d}$ to build ${\mathbf{R}}_{d}$, and subsequently ${\mathbf{q}_d}$.

To generate $\boldsymbol{\omega}_d$ and $\boldsymbol{\alpha}_d$, we use $\dot{\mathbf{R}}_{d} =[\dot{\mathbf{b}}_{1_d},\dot{\mathbf{b}}_{2_d},\dot{\mathbf{b}}_{3_d}]$ and $\ddot{\mathbf{R}}_{d} = [\ddot{\mathbf{b}}_{1_d}, \ddot{\mathbf{b}}_{2_d}, \ddot{\mathbf{b}}_{3_d}]$. According to \cite{lee2010geometric}, we have
\begin{equation}
    \dot{\mathbf{b}}_{3_d} = \frac{\dot{\boldsymbol{\kappa}}}{\lVert \boldsymbol{\kappa} \rVert} - \frac{\boldsymbol{\kappa} \cdot \dot{\boldsymbol{\kappa}}}{\lVert \boldsymbol{\kappa} \rVert ^3} \boldsymbol{\kappa},
\end{equation}
\begin{equation}
    \dot{\boldsymbol{\nu}} = \dot{\mathbf{b}}_{3_d} \times \mathbf{b}_{1_r} + \mathbf{b}_{3_d} \times \dot{\mathbf{b}}_{1_r},
\end{equation}
\begin{equation}
    \dot{\mathbf{b}}_{2_d} = \frac{\dot{\boldsymbol{\nu}}}{\lVert \boldsymbol{\nu} \rVert} - \frac{\boldsymbol{\nu} \cdot \dot{\boldsymbol{\nu}}}{\lVert \boldsymbol{\nu} \rVert ^3} \boldsymbol{\nu},
\end{equation}
\begin{equation}
    \dot{\mathbf{b}}_{1_d} = \dot{\mathbf{b}}_{2_d} \times \mathbf{b}_{3_d} + \mathbf{b}_{2_d} \times \dot{\mathbf{b}}_{3_d},
\end{equation}
\begin{equation}
\begin{split}
    \ddot{\mathbf{b}}_{3_d} = \frac{\ddot{\boldsymbol{\kappa}}}{\lVert \boldsymbol{\kappa} \rVert} - 2\frac{ \boldsymbol{\kappa} \cdot \dot{\boldsymbol{\kappa}}}{\lVert \boldsymbol{\kappa} \rVert ^3} \dot{\boldsymbol{\kappa}} - \frac{ \lVert \dot{\boldsymbol{\kappa}} \rVert^2 + \boldsymbol{\kappa} \cdot \ddot{\boldsymbol{\kappa}}}{\lVert \boldsymbol{\kappa} \rVert ^3} \boldsymbol{\kappa} \\
     + 3\frac{ (\boldsymbol{\kappa} \cdot \dot{\boldsymbol{\kappa}})^2}{\lVert \boldsymbol{\kappa} \rVert ^5} \boldsymbol{\kappa},
\end{split}
\end{equation}
\begin{equation}
    \ddot{\boldsymbol{\nu}} = {\ddot{\mathbf{b}}_{3_d} \times \mathbf{b}_{1_r} + \mathbf{b}_{3_d} \times \ddot{\mathbf{b}}_{1_d} + 2 \dot{\mathbf{b}}_{3_d} \times \dot{\mathbf{b}}_{1_d}},
\end{equation}
\begin{equation}
\begin{split}
    \ddot{\mathbf{b}}_{2_d} = \frac{\ddot{\boldsymbol{\nu}}}{\lVert \boldsymbol{\nu} \rVert} - 2\frac{ \boldsymbol{\nu} \cdot \dot{\boldsymbol{\nu}}}{\lVert \boldsymbol{\nu} \rVert ^3} \dot{\boldsymbol{\nu}} - \frac{ \lVert \dot{\boldsymbol{\nu}} \rVert^2 + \boldsymbol{\nu} \cdot \ddot{\boldsymbol{\nu}}}{\lVert \boldsymbol{\nu} \rVert ^3} \boldsymbol{\nu} \\
     + 3\frac{ (\boldsymbol{\nu} \cdot \dot{\boldsymbol{\nu}})^2}{\lVert \boldsymbol{\nu} \rVert ^5} \boldsymbol{\nu},
\end{split}
\end{equation}
and
\begin{equation}
    \ddot{\mathbf{b}}_{1_d} = {\ddot{\mathbf{b}}_{2_d} \times \mathbf{b}_{3_d} + \mathbf{b}_{2_d} \times \ddot{\mathbf{b}}_{3_d} + 2 \dot{\mathbf{b}}_{2_d} \times \dot{\mathbf{b}}_{3_d}}.
\end{equation}
Using \eqref{eq:A}, we can compute $\dot{\boldsymbol{\kappa}}$ and $\ddot{\boldsymbol{\kappa}}$ as follows
\begin{equation}
    \dot{\boldsymbol{\kappa}} =  {m \left(\dddot{\boldsymbol{\xi}}_d -\operatorname{diag}(\boldsymbol{\lambda}_{\xi}^{\top})\mathbf{a}_{e}  -\mathbf{K}_{\xi}\sech(\mathbf{s_{\xi}})^{2} \circ \dot{\mathbf{s}}_{\xi}\right)},
\end{equation}
\begin{equation}
\begin{split}
    \ddot{\boldsymbol{\kappa}} =  m \Bigl( \boldsymbol{\xi}^{(4)}_d -\operatorname{diag}(\boldsymbol{\lambda}_{\xi}^{\top}) \mathbf{j}_{e}  - \mathbf{K}_{\xi}\sech(\mathbf{s}_{\xi})^{2} \circ \ddot{\mathbf{s}}_{\xi}+\Bigr. \\ \Bigl. 2\mathbf{K}_{\xi}\sech(\mathbf{s}_{\xi})^{2} \circ \tanh(\mathbf{s_{\xi}}) \circ \dot{\mathbf{s}}^{\circ 2}_{\xi}\Bigr),
    \end{split}
\end{equation}
where $\mathbf{a}_{e}$ and $\mathbf{j}_{e}$ are the acceleration and jerk errors, respectively. We can now generate $\boldsymbol{\omega}_d$ and $\boldsymbol{\alpha}_d$ as follows
\begin{equation}
    \boldsymbol{\omega}_d = (\mathbf{R}_d^{\top} \dot{\mathbf{R}}_d)^{\vee},\;\text{and}\;
    \boldsymbol{\alpha}_d = (\mathbf{R}_d^{\top} \ddot{\mathbf{R}}_d-{\hat{\boldsymbol{\omega}}_d}^2 )^{\vee},
\end{equation}
where $^\vee:\mathfrak{so}(3)\rightarrow \mathbb{R}^3$ and $\hat{\cdot}:\mathbb{R}^3\rightarrow \mathfrak{so}(3)$.






\subsection{Attitude controller}\label{subsec:attControl}
For the attitude controller, we adopt the method proposed in \cite{lopez2020sliding}.
The advantage of this approach is capturing the geometry of $\mathbb{S}^3$ and global stability as opposed to common geometric methods that are almost globally stable \cite{bhat2000topological}.
To this end, let us define the quaternion and angular velocity errors as
\begin{equation}\label{eq:quat_error}
    \mathbf{q}_e = \mathbf{q}_d^* \otimes \mathbf{q},\;\text{and}\;
    \boldsymbol{\omega}_e = \boldsymbol{\omega} - \boldsymbol{\omega}_d.
\end{equation}
The quaternion error dynamics becomes
\begin{equation}
    \dot{\mathbf{q}}_e = \frac{1}{2}\mathbf{q}_e \otimes \begin{bmatrix}
        0\\\boldsymbol{\omega}_e 
    \end{bmatrix}=\frac{1}{2}\begin{bmatrix}
            -\Vec{\mathbf{q}}_e \cdot \boldsymbol{\omega}_e\\ q_{w_e}\boldsymbol{\omega}_e+\Vec{\mathbf{q}}_e \times \boldsymbol{\omega}_e
        \end{bmatrix}.
\end{equation}
Let us define the attitude controller sliding surface as follows
\begin{equation}
    \mathbf{s}_q = \boldsymbol{\omega}_{e} + \operatorname{diag}(\boldsymbol{\lambda}_{q}^{\top})  \operatorname{sgn_{+}}(q_{w_e})\Vec{\mathbf{q}}_e,
\end{equation}
where
\begin{equation}
    \operatorname{sgn}_{+}(\cdot) = \left\{\begin{array}{cc}
 1 & \text{if } \cdot \geq 0,\\
 -1 & \text{if } \cdot < 0,
\end{array}\right.
\end{equation}
and $\boldsymbol{\lambda}_{q}$ is a design parameter with positive elements.
The term $\operatorname{sgn_{+}}(q_{w_e})$ is the key in this design. It maintains the control input continuity, thus eliminating the unwinding behavior of quaternion-based controllers. Next, according to \cite{lopez2020sliding}, the attitude control law becomes
\begin{equation}
\begin{split}
    \boldsymbol{\tau} = \mathbf{J}\boldsymbol{\alpha}_{d} + \boldsymbol{\omega} \times \mathbf{J}\boldsymbol{\omega} -\mathbf{J}  \operatorname{diag}(\boldsymbol{\lambda}_{q}^{\top}) \operatorname{sgn_{+}}(q_{w_e})\dot{\Vec{\mathbf{q}}}_e -\\
    \mathbf{K}_q \tanh(\mathbf{s}_q),
    \end{split}
\end{equation}
where $\mathbf{K}_q$ is a design parameter.
\subsection{Stability analysis}\label{subsec:LyapunovProof}
The global stability of the system can be shown using the following Lyapunov function
\begin{equation}
    V = \frac{1}{2} \mathbf{s^{\top}_{\xi}}\mathbf{s_{\xi}} + \frac{1}{2} \mathbf{s}_q^{\top} \mathbf{J} \mathbf{s}_q,
\end{equation}
provided that the design parameters $\mathbf{K}_{\xi}$ and $\mathbf{K}_q$ are sufficiently large. 
The steps to show the stability are straightforward using standard SMC stability analysis \cite{slotine1991applied}, and the mathematical treatment presented in \cite{lopez2020sliding}.
To avoid repetitions, and use the limited space to present our results, we skip the details of this part.





\section{RESULTS}
This section presents our simulation results.
We compare our method with two common 6-DOF flight control techniques, the geometric controller \cite{lee2010geometric}, and the Euler-based sliding mode controller \cite{nadda_adaptive_2018}. 
Also, to highlight the advantages of our attitude controller, we compare our results with a commonly used quaternion-based proportional-derivative (PD) attitude controller \cite{fresk2013full} combined with our position controller.
The simulation scenarios consist of (1) flip-maneuver, and (2) lemniscate trajectory tracking, both in the presence of parametric uncertainties and external disturbances.
Table \ref{tab:params} gives the parameter values used in simulations. 
Note that we meticulously fine-tuned all controllers over many trials to ensure their best performance.
\begin{table}[t]
\caption{Parameters values used in simulations}
\label{tab:params}
\centering
\begin{tabular}{@{}lll@{}}
\toprule
& Parameter  & Value \\ \midrule
\multicolumn{3}{l}{\textit{a) Vehicle parameters:}} \\
& $m$ & $27\;[g]$ \\
& ${\bf{J}} $ & ${\rm{diag}}\left( {1.66,1.66,2.93} \right) \times {10^{ - 5}}\;[kg{m^2}] $\\
& $c_t$ & $ 2.88 \times {10^{ - 8}}\;[\frac{N}{s^2}] $ \\
& $c_q$ & $ 7.24 \times {10^{ - 10}}\;[\frac{Nm}{s^2}] $ \\
& $l$ & $92\;[mm]$ \\
& $\beta$ & $45^{\circ}$ \\
& $\mathbf{d_a}$ & $2 \sin(\pi t + \frac{\pi}{2})\;[\frac{m}{s^2}]$ \\
& $\mathbf{d_{\alpha}}$ & $\sin(\pi t)\;[\frac{rad}{s^2}]$ \\
& $f^i_{min},f^i_{max} $ & $0.01, 0.15\;[N]\; where\; i=1,2,3,4$ \\
\multicolumn{3}{l}{\textit{b) Parameters common among all controllers:}} \\
& $m$ & $21.6\;[g]$ \\
& ${\bf{J}} $ & ${\rm{diag}}\left( {1.992,1.4940,3.0765} \right) \times {10^{ - 5}}\;[kg{m^2}] $\\
\multicolumn{3}{l}{\textit{c) Our controller parameters:}} \\\hspace{1em}
 & ${{\boldsymbol{\lambda}}_{\xi}} $ & $ \left[2, 4, 8\right]^{\top}$ \\
 & ${{\bf{K}}_{\xi}}$ & $\operatorname{diag}(\left[4, 2, 8\right])$ \\
 & ${{\boldsymbol{\lambda}}_{q}}$ & $\left[20, 20, 20\right]^{\top}$ \\
 & ${{\bf{K}}_{q}}$ & $ \operatorname{diag}([0.02, 0.02, 0.02])$ \\
\multicolumn{3}{l}{\textit{d) Geometric controller parameters \cite{lee2010geometric}:}} \\
  & ${{\bf{K}}_{\xi}}$ & $\operatorname{diag}(\left[5, 5, 15\right]\times m)$ \\
  & ${{\bf{K}}_{v}}$ & $\operatorname{diag}(\left[1, 1, 5\right]\times m)$ \\
  & ${{\bf{K}}_{j}}$ & $\operatorname{diag}(\left[0.01, 0.01, 0.01\right]\times m)$ \\
  & ${{\bf{K}}_{R}}$ & $\operatorname{diag}(\left[1.2, 0.5, 0.5\right])$ \\
  & ${{\bf{K}}_{\omega}}$ & $\operatorname{diag}(\left[0.02, 0.01, 0.01\right])$ \\
 \multicolumn{3}{l}{\textit{e) Euler-based SMC parameters \cite{nadda_adaptive_2018}:}} \\
 & ${{\boldsymbol{\lambda}}_{\xi}} $ & $ \left[2, 2, 2\right]^{\top}$ \\
 & ${{\bf{K}}_{\xi}}$ & $\operatorname{diag}(\left[4, 4, 30\right])$ \\
 & ${{\boldsymbol{\lambda}}_{\Phi}}$ & $\left[5,5,5\right]^{\top}$ \\
 & ${{\bf{K}}_{\Phi}}$ & $ \operatorname{diag}([20, 20, 20])$ \\
 \multicolumn{3}{l}{\textit{f) Quaternion-based PD controller parameters \cite{fresk2013full}:}} \\
 & ${{\boldsymbol{\lambda}}_{\xi}} $ & $ \left[2, 4, 8\right]^{\top}$ \\
 & ${{\bf{K}}_{\xi}}$ & $\operatorname{diag}(\left[4, 2, 8\right])$\\
   & ${{\bf{K}}_{q}}$ & $\operatorname{diag}(\left[0.05, 0.05, 0.05\right])$ \\
  & ${{\bf{K}}_{\omega}}$ & $\operatorname{diag}(\left[0.001, 0.001, 0.001\right])$ \\
 \bottomrule
\end{tabular}
\end{table}
\subsection{Scenario 1: flip over maneuver}
In this scenario, the vehicle starts from $\boldsymbol{\xi}_0 =\left[0,0,2\right]^{\top}$, $\mathbf{v}_0 = \left[1,1,3\right]^{\top}$, $\mathbf{q}_0 =\left[0,0,0,1\right]^{\top}$, and $\boldsymbol{\omega}_0 = \left[0,0,0\right]^{\top}$, conducting a flip maneuver to reach $\boldsymbol{\xi}_d =\left[1,2,3\right]^{\top}$, and $\mathbf{b}_{1_r} =\left[1,0,0\right]^{\top}$.
Note that the vehicle starts from an inverted orientation, with an initial linear velocity.
This represents a challenging aggressive deployment scenario.

\begin{figure}[t]
    \centering
    \includegraphics[width=0.93\linewidth]
{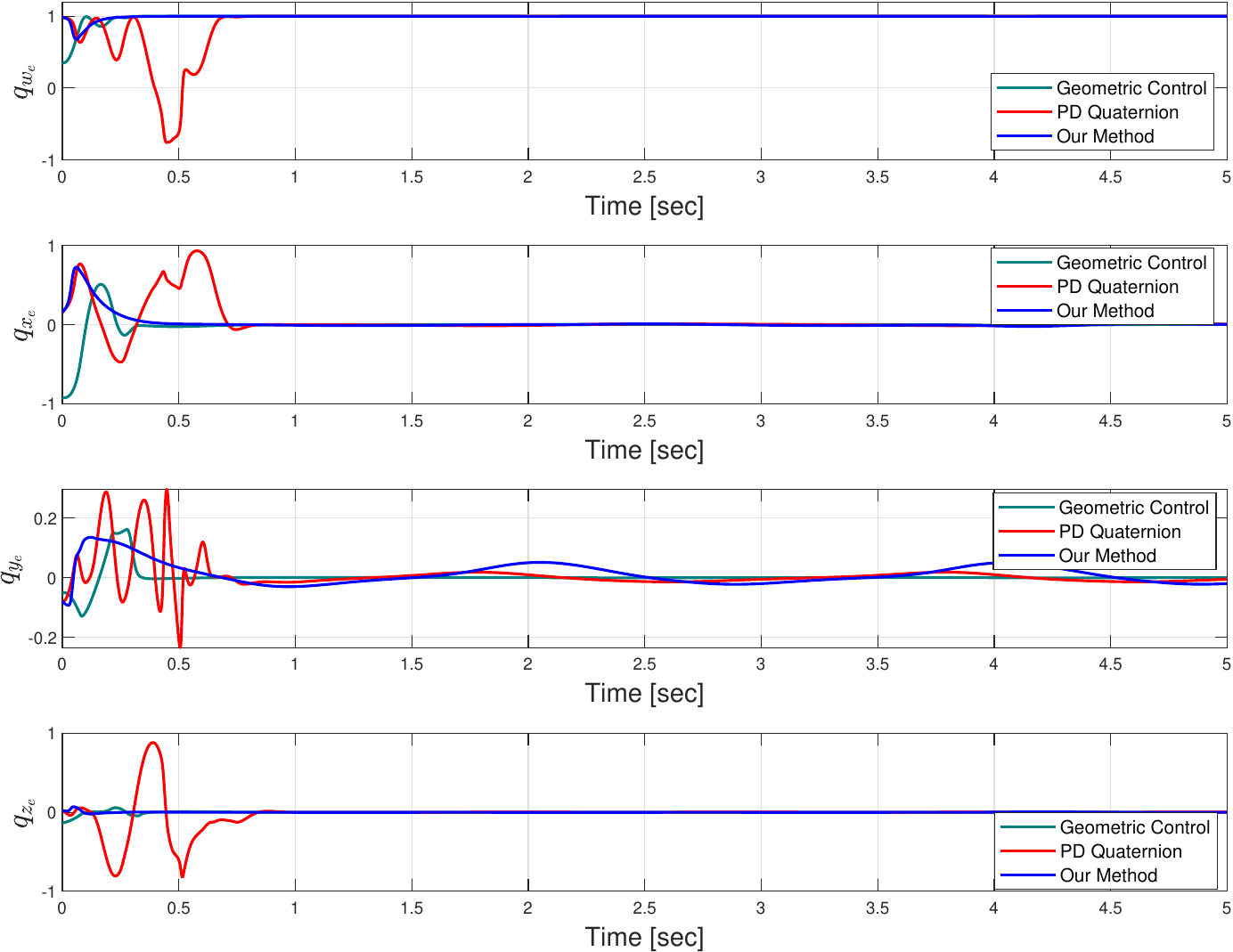}
    \caption{Quaternion error for flip-over maneuver scenario. The different initial values of the geometric controller are due to the difference in its reference generation method within its position controller.
}
    \label{fig:e_q_a}
    \centering
    \includegraphics[width=0.93\linewidth]{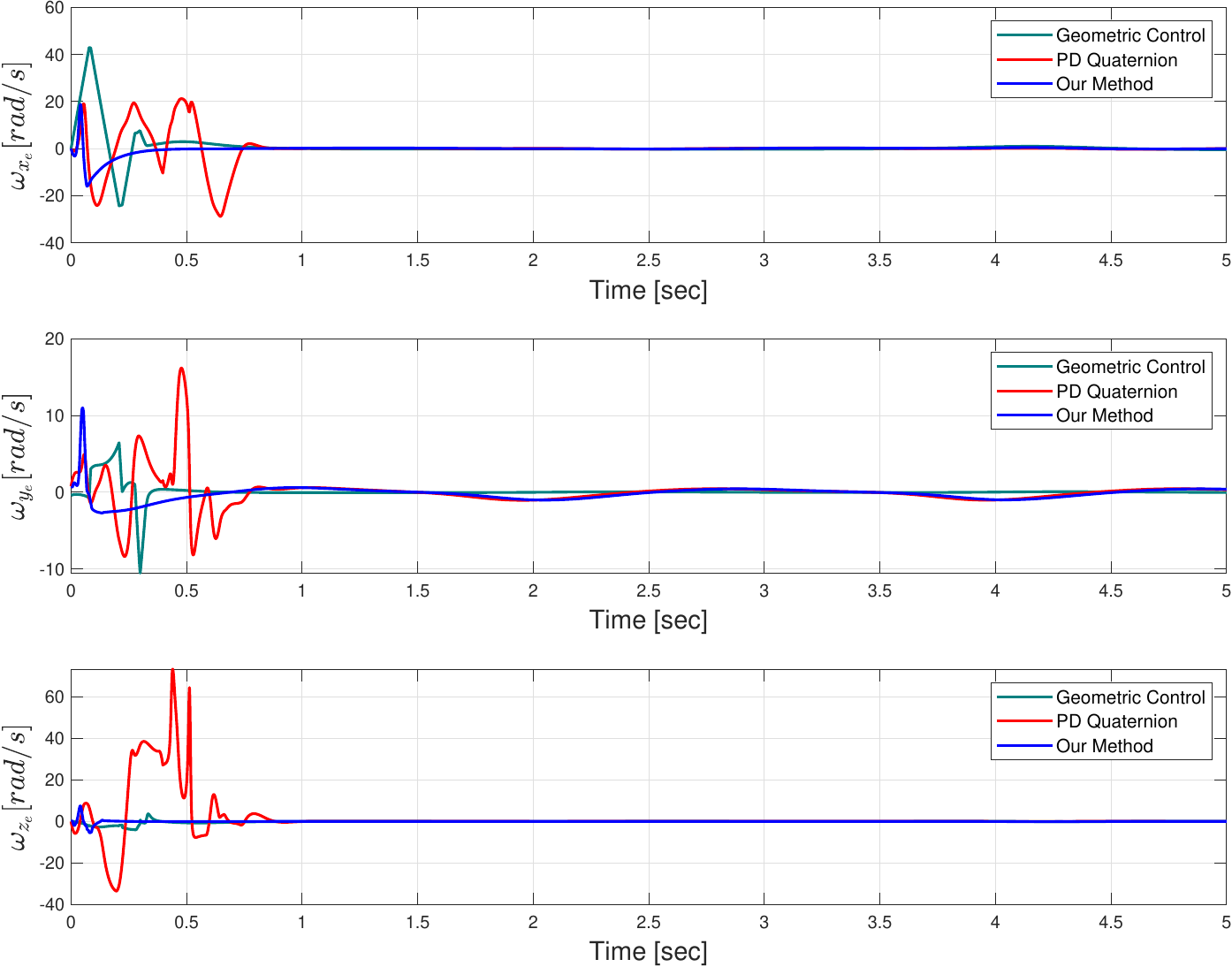}
    \caption{Angular velocity error in flip-over maneuver scenario.}
    \label{fig:e_omega_a}
\end{figure}

\begin{figure}[t]
    \centering
    \includegraphics[width=0.93\linewidth]{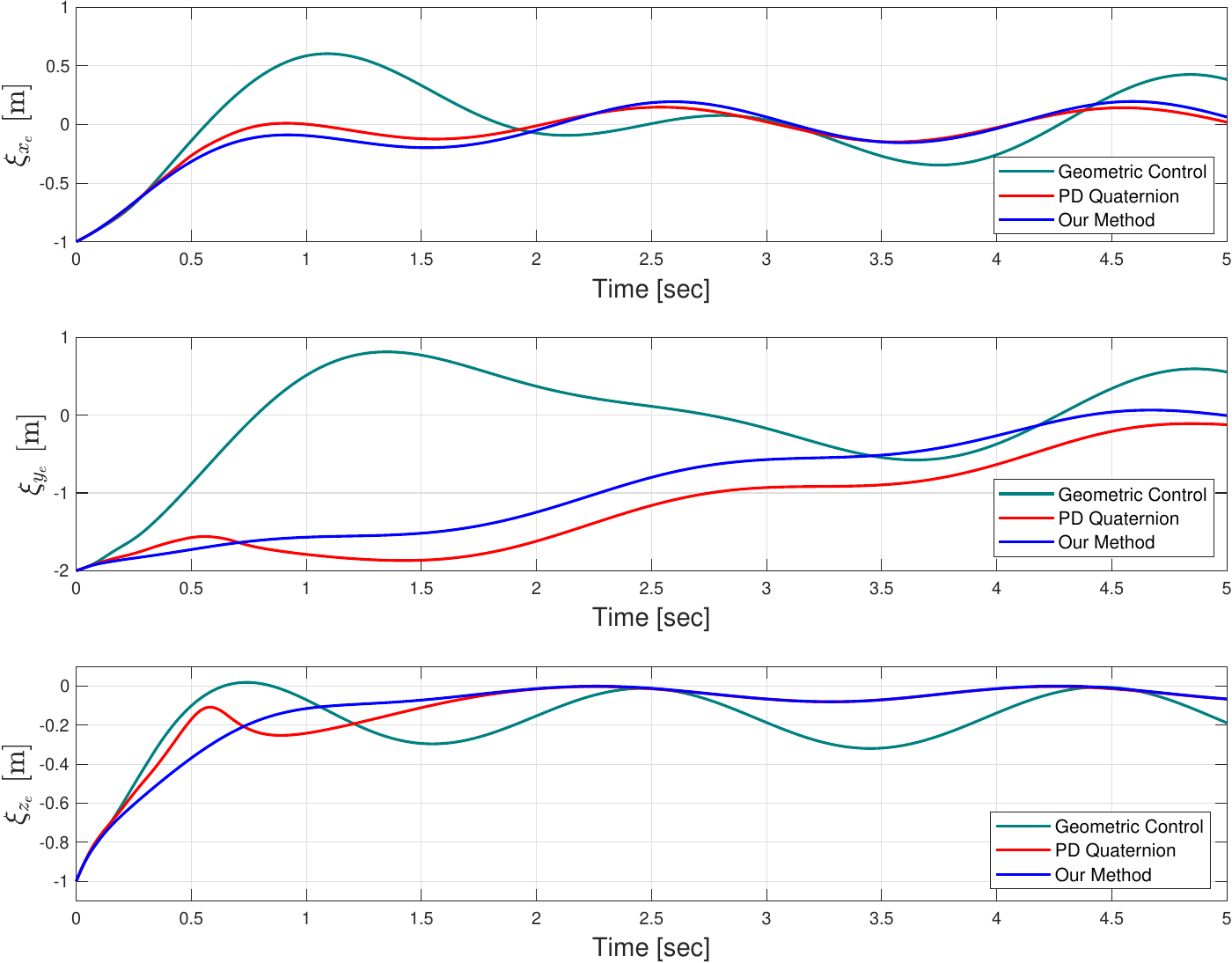}
    \caption{Position error in flip-over maneuver scenario.}
    \label{fig:e_p_a}
    \centering
    \includegraphics[width=0.93\linewidth]{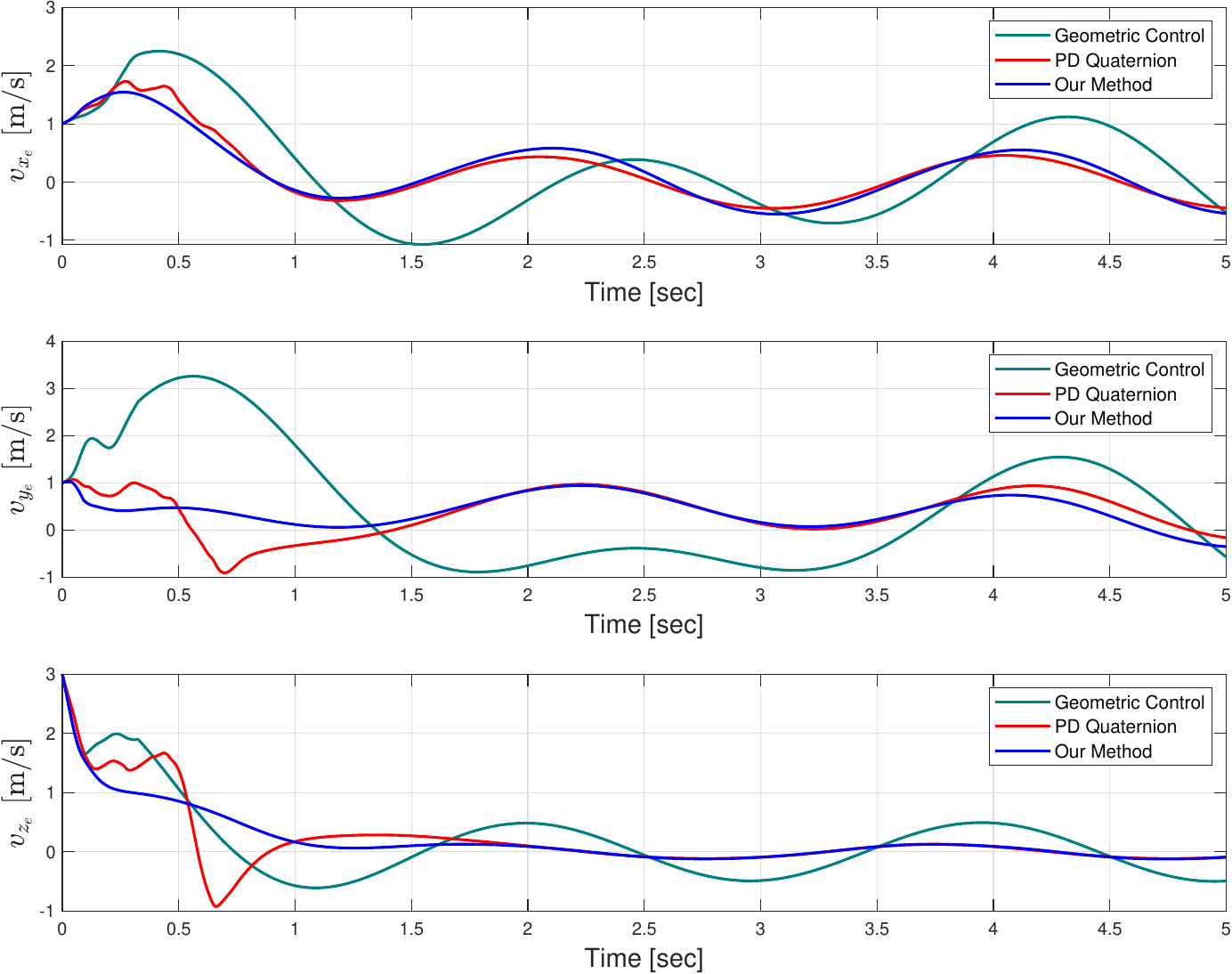}
    \caption{Linear velocity error in flip-over maneuver scenario.}
    \label{fig:e_v_a}
\end{figure}

\begin{figure}[t]
    \centering
        \includegraphics[width=0.93\linewidth]{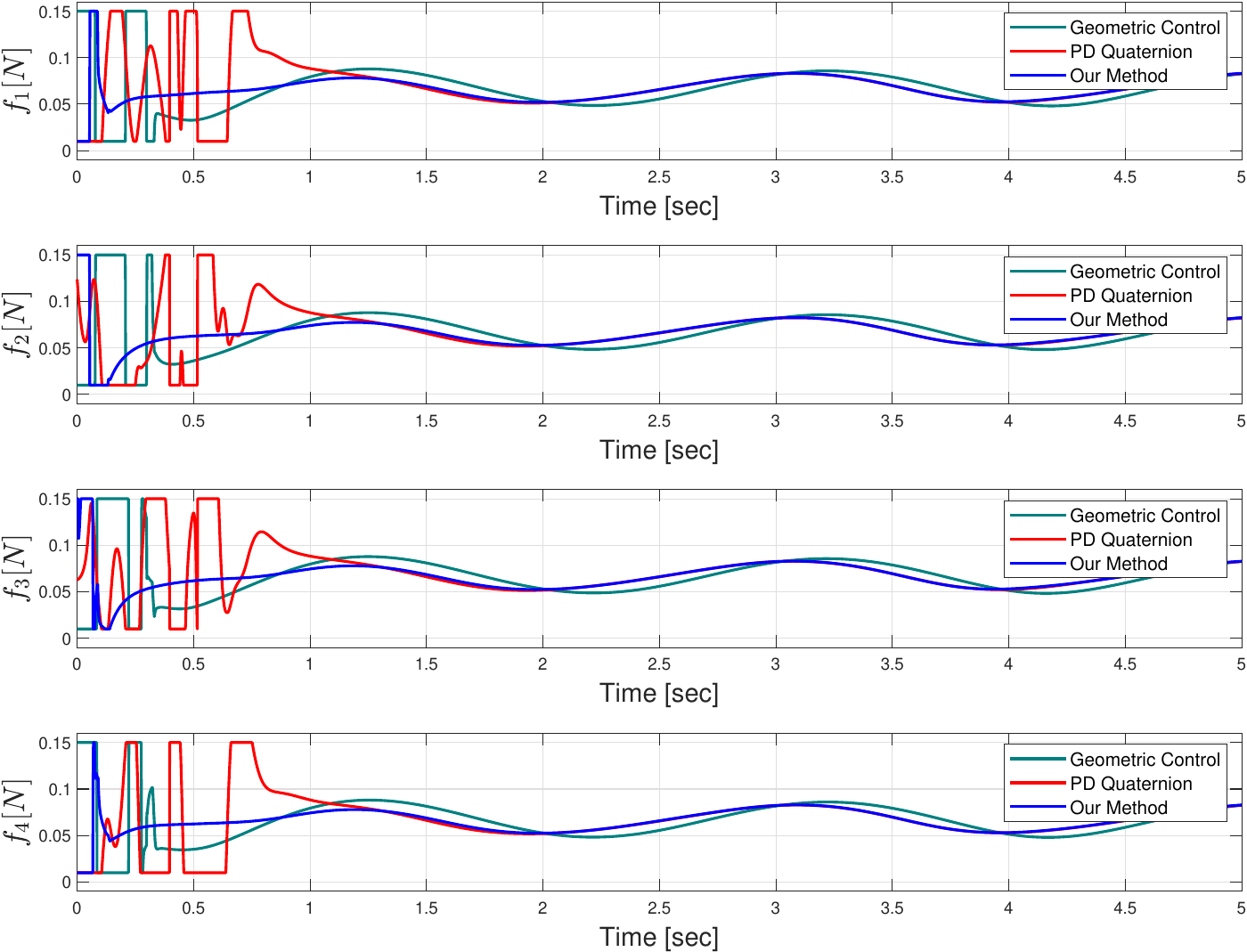}
    \caption{Thrust generated by each motor in the flip-over maneuver scenario.}
    \label{fig:f_a}
\end{figure}

Figures \ref{fig:e_q_a} - \ref{fig:f_a} illustrate the results.
The Euler-based SMC failed to perform this maneuver due to the singularity of Euler angles in the inverted orientation.
Even with initial conditions slightly away from the inverted orientation and extensive parameter tuning, achieving stable behavior with this method proved unsuccessful.
This stems from the $\dot{\boldsymbol{\eta}} \approx {\boldsymbol{\omega}}$ simplification that is invalid for large $\phi$ and $\theta$ values, a limitation of the previous 6-DOF SMC designs.
Our method and the two other benchmark methods successfully conducted this maneuver; however, the quaternion-based PD method suffers from high fluctuations in rotational error dynamics. 
Our method and the geometric controller exhibit comparable performance in rotational error dynamics; however, our method offers faster convergence and less steady-state error in transnational error dynamics.
Interestingly, this is achieved by a smaller control input and fewer actuator saturation, highlighting the efficiency of our method.

\subsection{Scenario 2: lemniscate trajectory tracking}
In this scenario, the vehicle starts from $\boldsymbol{\xi}_0 =\left[0,0,2\right]^{\top}$, $\mathbf{v}_0 =\left[1,-0.5,0.5\right]^{\top}$, $\mathbf{q}_0 =\left[0.2837,0,0,-0.9589\right]^{\top}$, and $\boldsymbol{\omega}_0 =\left[0,0,0\right]^{\top}$, tracking a lemniscate trajectory with varying speed and acceleration, reaching to maximum $2.51\;[m/s]$ and $1.7\;[m/s^2]$.
Note that the vehicle under consideration is a nano quadrotor and such speeds and accelerations are excessive for the vehicle.

\begin{figure}[t]
    \centering
    \includegraphics[width=0.93\linewidth]{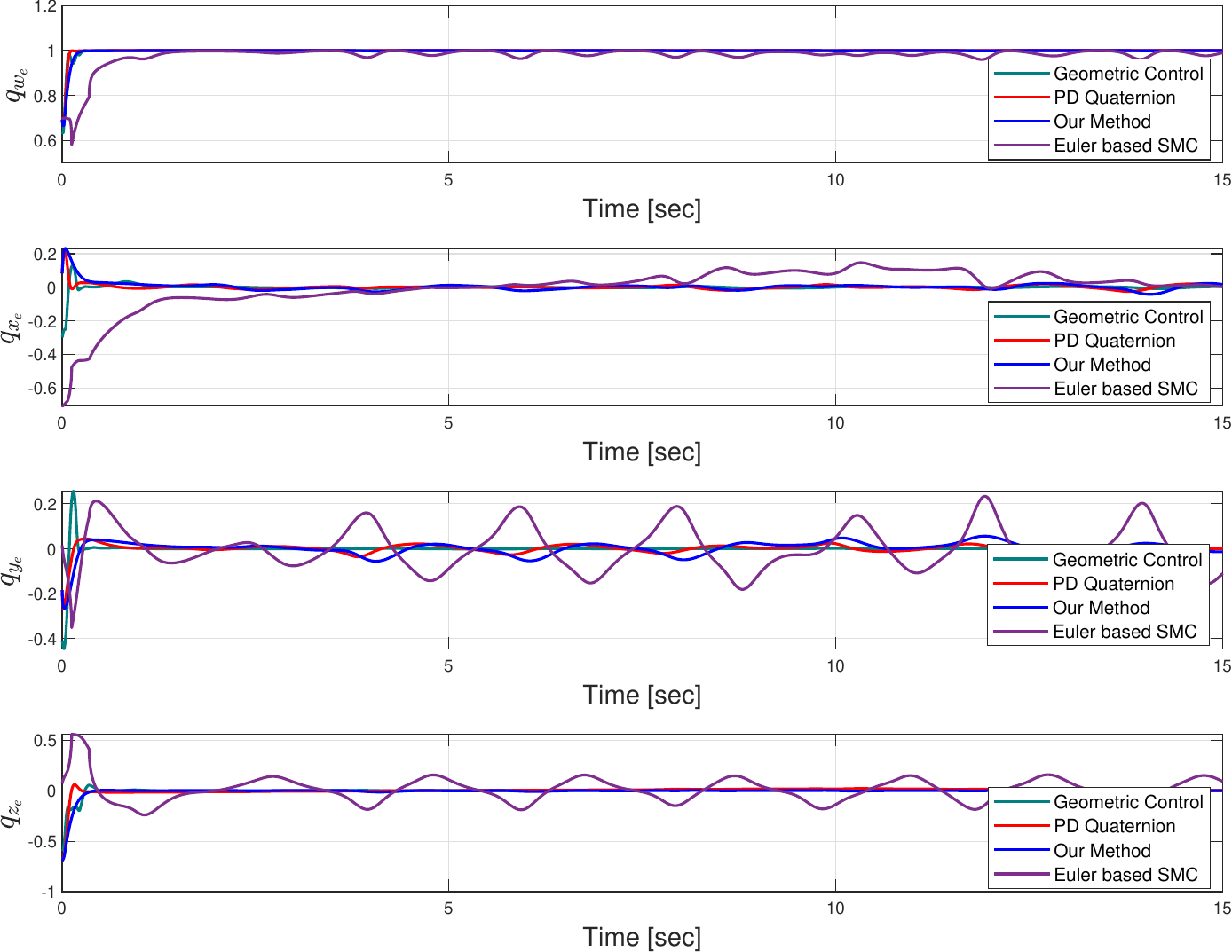}
    \caption{Quaternion error in the lemniscate trajectory tracking scenario.}
    \label{fig:e_q_l}
    \end{figure}

\begin{figure}[t]
    \centering
    \includegraphics[width=0.93\linewidth]{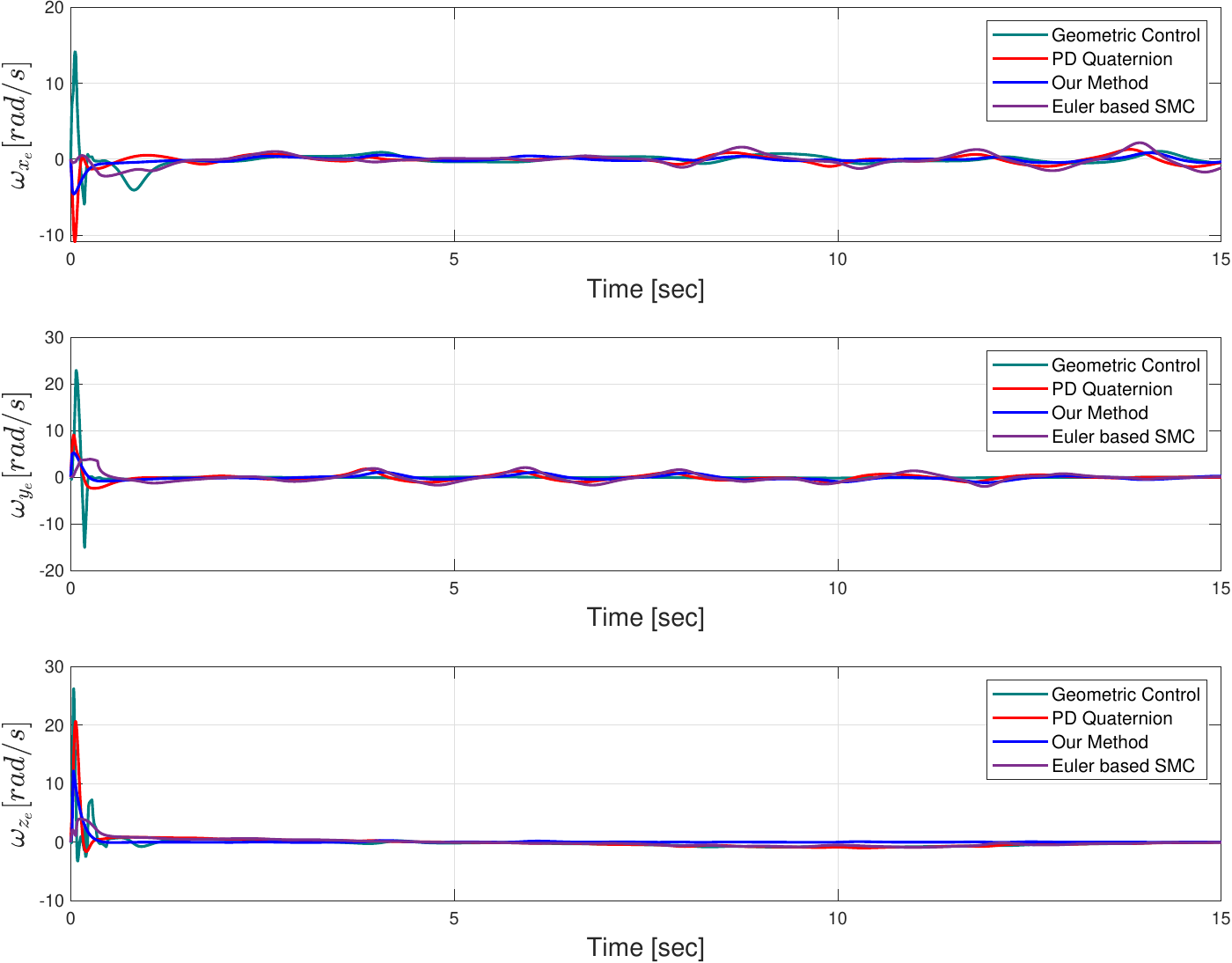}
    \caption{Angular velocity error in the lemniscate trajectory tracking scenario.}
    \label{fig:e_omega_l}
\end{figure}

\begin{figure}[t]
    \centering
    \includegraphics[width=0.93\linewidth]{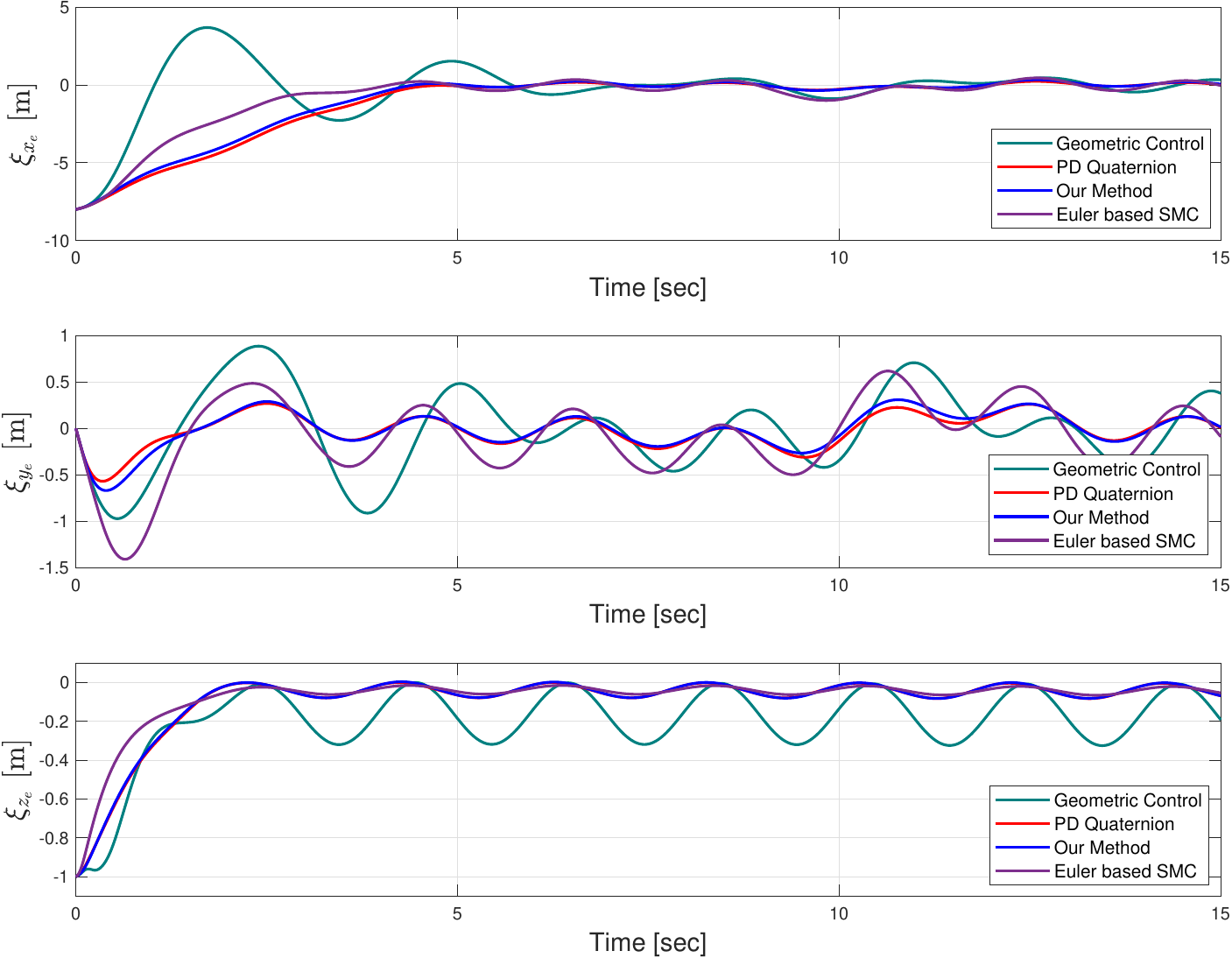}
    \caption{Position error in the lemniscate trajectory tracking scenario.}
    \label{fig:e_p_l}
    \end{figure}

\begin{figure}
    \centering
    \includegraphics[width=0.93\linewidth]{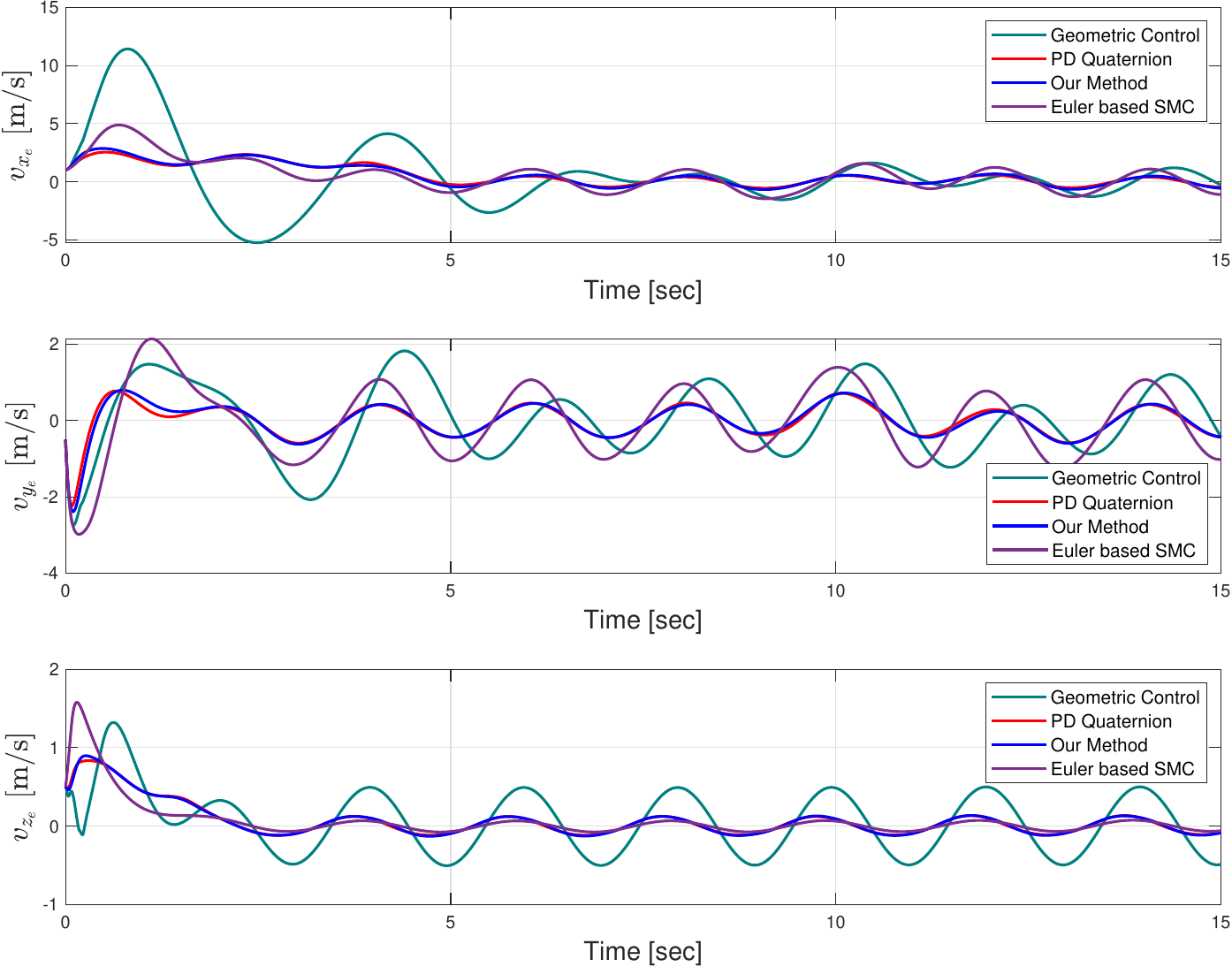}
    \caption{Linear velocity error in the lemniscate trajectory tracking scenario.}
    \label{e_v_l}
\end{figure}

\begin{figure}
    \centering
    \includegraphics[width=0.93\linewidth]{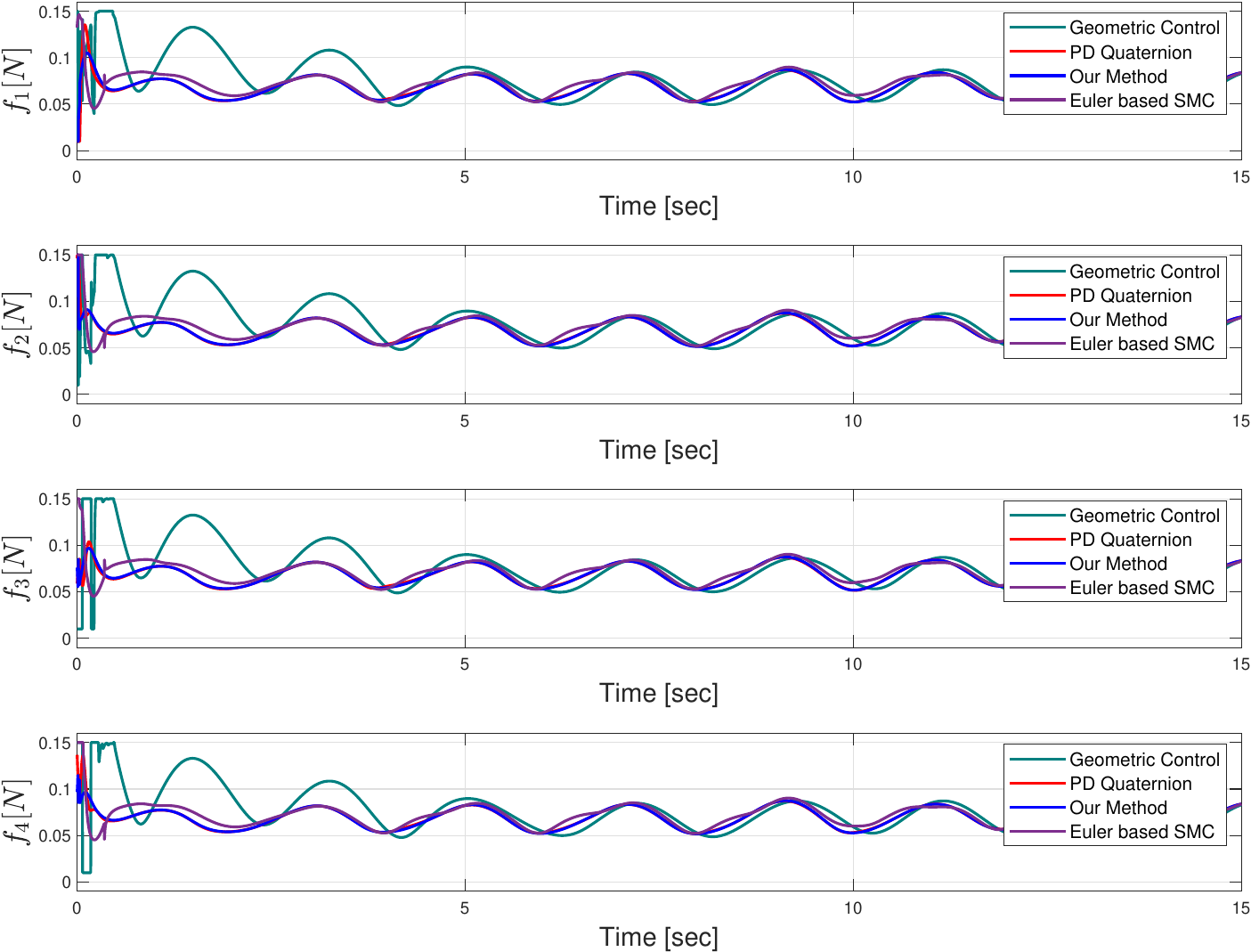}
    \caption{Thrust generated by each motor in the lemniscate trajectory tracking scenario.}
    \label{fig:f_l}
\end{figure}

Figures \ref{fig:e_q_l}-\ref{fig:f_l} illustrate the results.
Concerning rotational dynamics, the quaternion-based and geometric controllers manage to keep quaternion and angular velocity errors relatively small; however, the Euler-based SMC falls behind.
Due to the high acceleration and speed profile of this trajectory, the roll angles reach values up to $45^{\circ}$, detrimental to the validity of the simplified $\dot{\boldsymbol{\eta}} \approx {\boldsymbol{\omega}}$ model.
This may also explain why the Euler-based SMC exhibits slower convergence and larger steady-state errors in translational dynamics compared to the two other SMC controllers.
Interestingly, while our method and the quaternion-based PD controllers have identical position controllers, our method offers a slightly better performance which can be attributed to its more accurate attitude-tracking performance.
Note that all SMC position controllers outperform the geometric controller in translational dynamics with smaller control efforts and fewer actuator saturations.
This can be explained by the higher robustness offered by SMC compared to the PD-like position control in the geometric controller.
Overall, our proposed method offers the smallest control effort with the fastest error convergence and smallest steady-state error.

\section{Conclusion}
This paper presented a new 6-DOF sliding mode controller for quadrotors.
The strengths of our method include (1) global stability, (2) avoiding model oversimplification, and (3) eliminating the unwinding problem of quaternion-based controllers.
These features overcome the limitations of existing SMC designs for quadrotor flight control, enabling a fast and robust controller that outperforms several existing methods over challenging scenarios while minimizing control efforts and actuator saturations. 
Possible directions for future work include comparison with other agile flight controllers like differential flatness and nonlinear model predictive control methods, higher-order SMC extensions of the current formulation, and enhancing the controller with adaption laws.

\bibliographystyle{IEEEtran}
\bibliography{References}
\end{document}